\documentclass{ieeeaccess}
\usepackage{cite}
\usepackage{amsmath,amssymb,amsfonts}
\usepackage{algorithmic}
\usepackage{algorithm}
\usepackage{graphicx}
\usepackage{textcomp}
\usepackage{subfigure}
\newcommand{\eg}{{e.g.}}
\newcommand{\ie}{{i.e.}}
\newcommand{\etal}{\textit{et al.~}}

\def\BibTeX{{\rm B\kern-.05em{\sc i\kern-.025em b}\kern-.08em
    T\kern-.1667em\lower.7ex\hbox{E}\kern-.125emX}}
\begin{document}
\history{Date of publication xxxx 00, 0000, date of current version xxxx 00, 0000.}
\doi{10.1109/ACCESS.2017.DOI}

\title{Locality Constraint Dictionary Learning with Support Vector for Pattern Classification}
\author{\uppercase{He-Feng Yin}\authorrefmark{1,2},
\uppercase{Xiao-Jun Wu\authorrefmark{1,2}, and Su-Gen Chen}\authorrefmark{3}}
\address[1]{School of Internet of Things Engineering, Jiangnan University, Wuxi 214122, China}
\address[2]{Jiangsu Provincial Engineering Laboratory of Pattern Recognition and Computational Intelligence, Jiangnan University, Wuxi 214122, China}
\address[3]{School of Mathematics and Computational Science, Anqing Normal University, Anqing 246133, China}

\tfootnote{This work was supported by the National Natural Science Foundation of China (Grant No. 61672265, U1836218, 61702012), the 111 Project of Ministry of Education of China (Grant No. B12018), the Postgraduate Research \& Practice Innovation Program of Jiangsu Province under Grant No. KYLX\_1123, the Overseas Studies Program for Postgraduates of Jiangnan University and the China Scholarship Council (CSC, No.201706790096).}

\markboth
{H. Yin \headeretal: LCDL-SV for Pattern Classification}
{H. Yin \headeretal: LCDL-SV for Pattern Classification}

\corresp{Corresponding author: Xiao-Jun Wu (e-mail: wu\_xiaojun@jiangnan.edu.cn)}

\begin{abstract}
Discriminative dictionary learning (DDL) has recently gained significant attention due to its impressive performance in various pattern classification tasks. However, the locality of atoms is not fully explored in conventional DDL approaches which hampers their classification performance. In this paper, we propose a locality constraint dictionary learning with support vector discriminative term (LCDL-SV), in which the locality information is preserved by employing the graph Laplacian matrix of the learned dictionary. To jointly learn a classifier during the training phase, a support vector discriminative term is incorporated into the proposed objective function. Moreover, in the classification stage, the identity of test data is jointly determined by the regularized residual and the learned multi-class support vector machine. Finally, the resulting optimization problem is solved by utilizing the alternative strategy. Experimental results on benchmark databases demonstrate the superiority of our proposed method over previous dictionary learning approaches on both hand-crafted and deep features. \textit{The source code of our proposed LCDL-SV is accessible at https://github.com/yinhefeng/LCDL-SV}.
\end{abstract}

\begin{keywords}
Dictionary learning, support vector discriminative term, locality constraint, pattern classification
\end{keywords}

\titlepgskip=-15pt

\maketitle

\section{Introduction}
\label{sec:introduction}
Dictionary learning (DL) has aroused considerable interest during the past decade and has been adopted in a wide rang of applications, such as face recognition\cite{b1}, image fusion\cite{b2} and person re-identification\cite{b3,b3_1}. According to the characteristic of the learned dictionary, existing DL approaches for pattern classification can be divided into three categories: synthesis dictionary learning (SDL), analysis dictionary learning (ADL) and dictionary pair learning (DPL). In SDL, the dictionary is employed to represent the input data as a linear superposition of atoms. ADL aims to yield the sparse representation by exploiting the dictionary as a transformation matrix. DPL, also referred to as analysis-synthesis dictionary learning (ASDL), can jointly learn synthesis dictionary and analysis dictionary. According to whether the dictionary is class-shared or not, SDL can be further divided into three different types, \ie, shared SDL, class-specific SDL and hybrid SDL. Similarly, ADL can be classified into two categories, \ie, shared ADL and class-specific ADL. Fig. \ref{fig:taxonomy} presents a taxonomy of dictionary learning approaches for pattern classification.
\Figure[h!](topskip=0pt, botskip=0pt, midskip=0pt)[width=4 in]{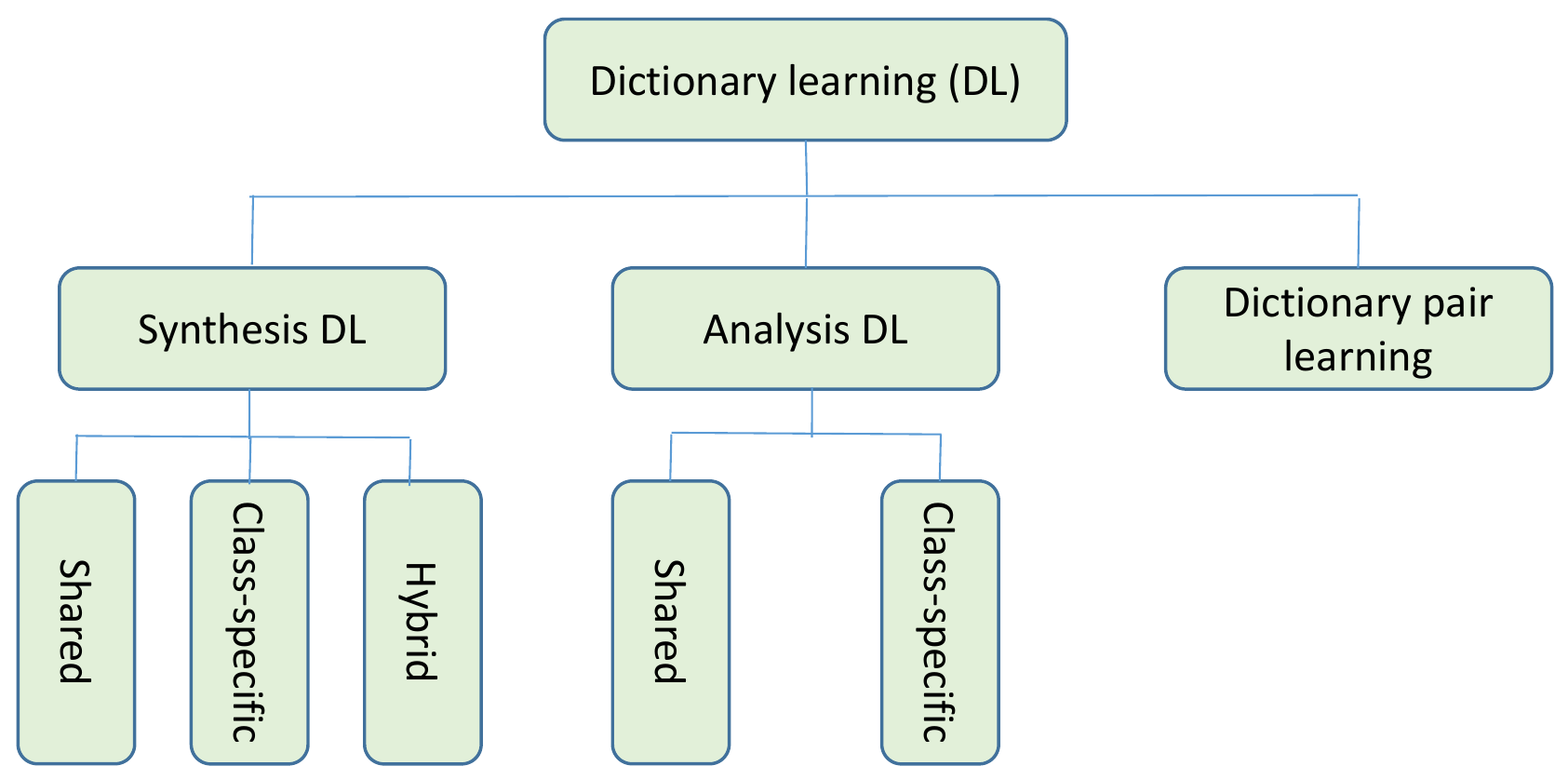}
{A taxonomy of dictionary learning approaches for pattern classification.\label{fig:taxonomy}}

In class-specific SDL, sub-dictionary for each class is independently learned, then all the sub-dictionaries are concatenated to form the final dictionary. Ramirez \etal \cite{b4} presented a dictionary learning with structured incoherence (DLSI) method by imposing incoherence constraint on sub-dictionaries so as to encourage dictionaries correspond to different classes to be as independent as possible. Yang \etal \cite{b5} proposed a metaface learning (MFL) algorithm which learns a set of metafaces for each class. Yang \etal \cite{b6} developed a Fisher discrimination dictionary learning (FDDL) method which imposes the Fisher discrimination criterion on the coding coefficients to learn class-specific sub-dictionaries. By considering the fact that different training samples contribute unequally to the dictionary, Liu \etal \cite{b7} proposed a class specific dictionary learning (CSDL) approach. Akhtar \etal \cite{b8} developed a joint discriminative Bayesian dictionary and classifier learning (JBDC) approach which associates the dictionary atoms with the class labels using Bernoulli distributions. By employing the directions of coefficients to promote the discriminative capability of representation, Wang \etal \cite{b9} presented a unidirectional representation dictionary learning (URDL) algorithm. Ling \etal \cite{b10} proposed a class-oriented discriminative DL (CODDL) method, in which the class-specific sub-dictionaries are learned in a classwise fashion.

In shared SDL, a universal dictionary shared by all classes is learned. The most classic SDL approach is the K-SVD algorithm \cite{b11} which has been successfully applied to image compression and denoising. However, KSVD mainly focuses on the representational ability of the dictionary without considering its capability for classification. To address this problem, Zhang \etal \cite{b12} proposed a discriminative K-SVD (D-KSVD) method by introducing the classification error into the framework of K-SVD. Jiang \etal \cite{b13} further incorporated a label consistency constraint into K-SVD and presented a label consistent K-SVD (LC-KSVD) algorithm. The $\ell_0$-norm sparse regularization term is used in LC-KSVD, which is difficult to find the optimum sparse solution. To overcome this limitation, Shao \etal \cite{b13_1} explored a label embedded dictionary learning (LEDL) method which utilizes the $\ell_1$-norm as the sparse regularization term. By jointly learning a multi-class support vector machine (SVM) classifier, Cai \etal \cite{b14} developed a support vector guided dictionary learning (SVGDL) model. Zhang \etal \cite{b15} designed class relatedness oriented discriminative DL (CRO-DDL) method which utilizes the $\ell_{1,\infty}$ norm constraint \cite{b16} on the coding coefficient matrix. By integrating multiple classifiers training into dictionary learning process, Quan \etal \cite{b17} presented a  multiple classifiers based dictionary learning (MCDL) method. Dong \etal \cite{b18} proposed an orthonormal DL method by exerting an orthonormal constraint on the learned dictionary to enforce the dictionary atoms to be as dissimilar as possible. Min \etal \cite{b18_1} constructed a Laplacian regularized locality-constrained coding (LapLLC) algorithm for image classification, in which the similarity matrix is defined on the training data. To fully exploit the locality and label information of the learned dictionary, Li \etal \cite{b19} constructed a locality-constrained and label embedding dictionary learning (LCLE-DL) algorithm. Song \etal \cite{b20} presented a class-wise discriminative DL (CW-DDL) method which introduces a label-aware constraint and graph regularization into the framework of SDL. By employing the profiles (row vectors of coding coefficient matrix) to construct discriminative terms in SDL, Li \etal \cite{b21} proposed an interactively constrained discriminative DL (IC-DDL) algorithm for image classification.

In hybrid SDL, a dictionary that contains several class-specific sub-dictionaries and a shared dictionary is learned. Kong \etal \cite{b22} proposed a DL approach dubbed DL-COPAR which explicitly learns the shared patterns (the commonality) and the class-specific dictionaries (the particularity). Gao \etal \cite{b23} developed a category-specific and shared dictionary learning (CSDL) method for fine-grained image categorization. Sun \etal \cite{b24} presented a discriminative group sparse dictionary learning (DGSDL) model which learns a class-specific sub-dictionary for each class as well as a common sub-dictionary shared by all classes. By introducing a cross-label suppression constraint and group regularization term into the framework of SDL, Wang \etal \cite{b25} designed a cross-label suppression discriminative DL (CLS-DDL) approach. Lin \etal \cite{b26} proposed a robust, discriminative and comprehensive dictionary learning (RDCDL) model which learns a class-shared dictionary, class-specific dictionaries and a disturbance dictionary to represent the commonality, particularity and disturbance components in the data. In order to tackle corrupted samples, Vu \etal \cite{b27} developed a low-rank shared dictionary learning (LRSDL) framework which simultaneously learns a set of common patterns and class-specific features for classification. By integrating the low-rank matrix recovery technique with the class-specific and class-shared dictionary learning, Rong \etal \cite{b28} explored a low-rank double dictionary learning (LR$\textrm{D}^2$L) approach. Du \etal \cite{b29} proposed a low-rank graph preserving discriminative dictionary learning (LRGPDDL) method which incorporates the low-rank constraint on the class-specific dictionaries, graph preserving criterion and the dictionary incoherence term into the framework of SDL. Readers can refer to \cite{b30} for a survey of SDL approaches.

Recently, ADL has received increasing attention due to its efficacy and efficiency, and shared ADL has been widely studied. Rubinstein \etal \cite{b31} presented analysis K-SVD which is parallel to the synthesis K-SVD\cite{b11}. Afterwards, Shekhar \etal \cite{b32} applied ADL to image classification tasks and obtained comparable or better recognition performance than conventional SDL models. To enhance the classification performance of ADL, Guo \etal \cite{b33} proposed discriminative ADL (DADL) method. By introducing a synthesis-linear-classifier-based error term into the basic ADL model, Wang \etal \cite{b34} presented a synthesis linear classifier based ADL (SLC-ADL) algorithm. By solving a joint learning of ADL and a linear classifier through K-SVD based technique, Wang \etal \cite{b35} designed a synthesis K-SVD based ADL (SK-SVDADL) method. Similar to LC-KSVD\cite{b13}, Tang \etal \cite{b36} incorporated the label consistency term and classification error term into the framework of ADL and developed a structured ADL (SADL) approach. Maggu \etal \cite{b37} proposed label consistent transform learning (LCTL) for hyperspectral image classification. In essence, transform learning and ADL have similar formulation. For class-specific ADL, Wang \etal \cite{b38} proposed a class-aware ADL model which learns a discriminative analysis sub-dictionary for each class.

In DPL, a pair of synthesis dictionary and analysis dictionary is learned from the input data. Gu \etal \cite{b39} presented a projective dictionary pair learning (PDPL) framework which jointly learns a synthesis dictionary and an analysis dictionary. To further enhance the discriminative ability of DPL, Chen \etal \cite{b40} developed a discriminative DL approach called DPL-SV which introduces a differentiable support vector discriminative term into the DPL model. DPL does not impose sparse constraint on the representation matrix, which may lose discriminative power of sparse property. To alleviate this problem, Zhang \etal \cite{b41} designed a joint label consistent embedding and dictionary learning (JEDL) model which explicitly exploit a sparse constraint on the representation matrix. To preserve the locality property of learned atoms in the synthesis dictionary, Zhang \etal \cite{b42} proposed a locality constrained projective dictionary learning (LC-PDL) method. By jointly learning a classifier with the dictionary pair, Yang \etal \cite{b43} explored a  discriminative analysis-synthesis dictionary learning (DASDL) model. To preserve the local geometry structure of input data, Chang \etal \cite{b44} presented a graph-regularized discriminative analysis-synthesis dictionary pair learning (GDASDL) model to enhance the classification performance of DASDL. To integrate structured dictionary learning, analysis representation and analysis classifier training into a unified framework, Zhang \etal \cite{b45} proposed an analysis discriminative dictionary learning (ADDL) algorithm. Inspired by the superiority of $\ell_{1,\infty}$ norm \cite{b16}, Wei \etal \cite{b46} developed a fast DDL (FaDDL) method for synthetic aperture radar (SAR) image classification. The ordinal locality of analysis dictionary is not fully exploited in the above DPL and its variant, to tackle this problem, Li \etal \cite{b47} proposed a discriminative low-rank analysis-synthesis dictionary learning (LR-ASDL) algorithm with the adaptively ordinal locality.

In addition to the above DL approaches, to deal with multi-view data, some multi-view DL methods have been presented recently. Wu \etal \cite{b47_1} offered a multi-view low-rank dictionary learning (MLDL) method for image classification. Wu \etal \cite{b47_2} proposed a multi-view discriminant dictionary learning via learning view-specific and shared structured dictionaries (MDVSD) for image classification, in which a structured dictionary shared by all views and multiple view-specific structured dictionaries are simultaneously learned. Ma \etal \cite{b47_3} developed a multi-view coupled dictionary pair learning (MVCDL) framework for person re-identification. Wu \etal \cite{b47_4} presented a multi-view synthesis and analysis dictionaries learning (MSADL) approach for pattern classification.

However, ADL often requires enormous atoms to achieve satisfactory results when applied to pattern classification. For hybrid SDL, how to choose the optimal number of shared atoms remains unresolved. Moreover, the optimization process of class-specific SDL is time-consuming, especially when the number of classes is large. In this paper, we propose a locality constraint dictionary learning with support vector discriminative term (LCDL-SV) for pattern classification, which belongs to the shared SDL category. A support vector discriminative term is introduced to promote the discrimination of coding coefficients. Since the original training data may contain noise or outliers, graph Laplacian matrix constructed on the original training samples cannot faithfully describe the manifold structure. To alleviate this problem, we employ a locality constraint on atoms. The atoms are updated in the dictionary learning procedure, thus the graph Laplacian matrix defined on the atoms is also updated. More importantly, to further enhance the classification performance of our proposed method, the regularized residual and the learned multi-class SVM classifier are jointly exploited to classify the test data. The flowchart of our proposed method for classification is illustrated in Fig. \ref{fig: flowcht}. Firstly, features are extracted from the original training and test samples, respectively. Then the training data is fed into our proposed dictionary learning algorithm, when the dictionary learning process is completed, a compact dictionary and multi-class SVM are obtained. Finally, the test data is classified based on the learned multi-class SVM and the regularized residual. Our main contributions are summarized as follows,
\begin{itemize}
\item A locality constraint of atoms is introduced in our approach, and this term can intrinsically inherit the manifold structure of training data.
\item In addition to the learned SVM classifier, we take the regularized residual into account to further promote the classification performance.
\item The resulting problem is solved elegantly by employing an alternative optimization technique.
\end{itemize}
The remainder of this paper is structured as follows. Section \ref{sec:sec2} reviews related work on SDL. In Section \ref{sec:sec3},  we present our proposed approach, and detailed optimization procedures are given in Section \ref{sec:sec4}. Section \ref{sec:sec5} reports experimental results on five benchmark datasets. Finally, conclusions are drawn in Section \ref{sec:sec6}.
\Figure[h!](topskip=0pt, botskip=0pt, midskip=0pt)[width=3 in]{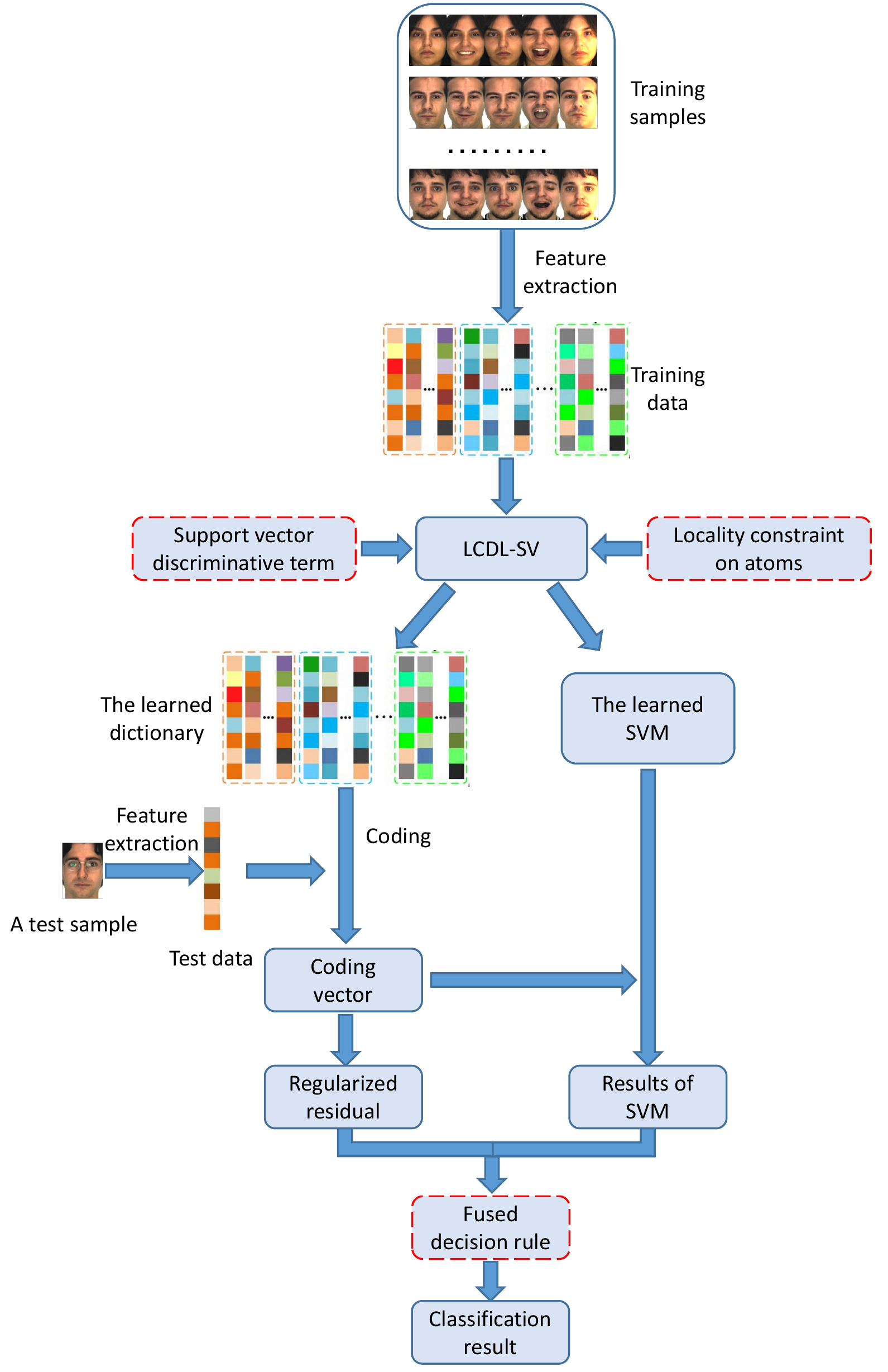}
{Flowchart of our proposed method for classification. In our LCDL-SV, a locality constraint on atoms and a support vector discriminative term are introduced in the formulation of LCDL-SV (red dashed rectangles in the middle part). Moreover, a fused decision strategy (red dashed rectangle at the bottom) which employs both the regularized residual and the learned multi-class SVM is designed to classify the input test sample.\label{fig: flowcht}}

\section{Related Work}
\label{sec:sec2}
In this section, we will briefly review some related work, including the basic K-SVD \cite{b11} and its two discriminative variants, \ie, D-KSVD \cite{b12} and LC-KSVD\cite{b13}. Additionally, support vector guided dictionary learning (SVGDL) method \cite{b14} is also introduced. To begin with, we first give an introduction to the notations used throughout this paper. Let $\mathbf{X}=[\mathbf{X}_1,\mathbf{X}_2,\ldots,\mathbf{X}_C]=[\boldsymbol{x}_1,\boldsymbol{x}_2,\ldots,\boldsymbol{x}_n]\in\mathbb{R}^{m\times n}$ be the data matrix of $n$ training samples belonging to $C$ classes, where $m$ is the dimension of vectorized data and $n$ is the total number of training samples, $\mathbf{D}=[\boldsymbol{d}_1,\boldsymbol{d}_2,\ldots,\boldsymbol{d}_K]\in\mathbb{R}^{m\times K}$ is the learned dictionary which has $K$ atoms, $\mathbf{Z}=[\boldsymbol{z}_1,\boldsymbol{z}_2,\ldots,\boldsymbol{z}_n]\in\mathbb{R}^{K\times n}$ is the coding coefficients matrix of $\mathbf{X}$ on the dictionary $\mathbf{D}$.

\subsection{K-SVD and its discriminative variants}
By generalizing the K-means clustering process, Aharon \etal \cite{b11} developed K-SVD to learn an overcomplete dictionary that best suits given data. The objective function of K-SVD is formulated as follows,
\begin{equation}
\label{eq:ksvd}
\underset{\mathbf{D},\mathbf{Z}}{\textrm{min}}\left \| \mathbf{X}-\mathbf{D}\mathbf{Z} \right \|_F^2, \ \textrm{s.t.} \ \left \| \boldsymbol{z}_i \right \|_0\leq T_0
\end{equation}
where $\mathbf{D}\in\mathbb{R}^{m\times K}$ is the dictionary that is to be learned, $\mathbf{Z}\in\mathbb{R}^{K\times n}$ is the coding coefficient matrix, and $T_0$ is a given sparsity level. (\ref{eq:ksvd}) can be solved by alternatively updating $\mathbf{D}$ and $\mathbf{Z}$. 
Although K-SVD yields impressive results in image compression and denoising, it is not tailored for classification. To make K-SVD suitable for classification problems, Zhang \etal \cite{b12} proposed D-KSVD algorithm by introducing the classification error term into the framework of K-SVD, 
\begin{equation}
\begin{split}
\label{eq:dksvd}
&\underset{\mathbf{D},\mathbf{W},\mathbf{Z}}{\textrm{min}}\left \| \mathbf{X}-\mathbf{D}\mathbf{Z} \right \|_F^2+\beta\left \| \mathbf{H}-\mathbf{W}\mathbf{Z} \right \|_F^2+\lambda\left \| \mathbf{W} \right \|_F^2,\\ & \textrm{s.t.} \ \left \| \boldsymbol{z}_i \right \|_0\leq T_0
\end{split}
\end{equation}
where $\mathbf{H}=[\boldsymbol{h}_1,\boldsymbol{h}_2,\ldots,\boldsymbol{h}_n]\in\mathbb{R}^{C\times n}$ is the label matrix of training data, $\boldsymbol{h}_i=[0,0,\ldots,1,\ldots,0,0]^T\in\mathbb{R}^{C\times 1}$ is the label vector of $\boldsymbol{x}_i$, and $\mathbf{W}$ is the parameters for a linear classifier. As can be seen from (\ref{eq:dksvd}), dictionary and a linear classifier are jointly learned in D-KSVD. Afterwards, Jiang \etal \cite{b13} presented LC-KSVD by solving the following optimization problem,
\begin{equation}
\begin{split}
\label{eq:lcksvd}
&\underset{\mathbf{D},\mathbf{W},\mathbf{A},\mathbf{Z}}{\textrm{min}}\left \| \mathbf{X}-\mathbf{D}\mathbf{Z} \right \|_F^2+\alpha\left \| \mathbf{Q}-\mathbf{A}\mathbf{Z} \right \|_F^2+\\&\beta\left \| \mathbf{H}-\mathbf{W}\mathbf{Z} \right \|_F^2, \ \textrm{s.t.} \ \left \| \boldsymbol{z}_i \right \|_0\leq T_0
\end{split}
\end{equation}
where $\mathbf{Q}=[\boldsymbol{q}_1,\boldsymbol{q}_2,\ldots,\boldsymbol{q}_n]\in\mathbb{R}^{K\times n}$ is an ideal representation matrix and $\mathbf{A}$ is a linear transformation matrix.

\subsection{SVGDL}
To promote the discriminative ability of coding vectors, Cai \etal \cite{b14} introduced a multi-class SVM regularization term into the framework of SDL. The regularization term is defined as follows,
\begin{equation}
\label{eq:svm_term}
L(\mathbf{Z})=2\sum_{c=1}^{C}f(\mathbf{Z},\boldsymbol{y}^c,\boldsymbol{u}_c,b_c)
\end{equation}
where $\boldsymbol{u}_c$ is the normal vector associated with the $c$-th class hyperplane of SVM, $b_c$ is the corresponding bias, and $\boldsymbol{y}^c=[y_1^c,y_2^c,\ldots,y_n^c]$ is defined as $y_i^c=1$ if class labels $y_i=c$ and otherwise $y_i^c=-1$. Concretely, the discrimination term is 
$f(\mathbf{Z},\boldsymbol{y},\boldsymbol{u},b)=\left \| \boldsymbol{u} \right \|_2^2+\theta\sum_{i=1}^{n}l(\boldsymbol{z}_i,\boldsymbol{y}_i,\boldsymbol{u},b)$, where $l(\boldsymbol{z}_i,\boldsymbol{y}_i,\boldsymbol{u},b)$ is the hinge loss function, and $\theta$ is a penalty parameter.

The objective function of SVGDL is formulated as follows,
\begin{equation}
\label{eq:svgdl}
\begin{split}
&\underset{\mathbf{D},\mathbf{Z},\mathbf{U},\boldsymbol{b}}{\textrm{min}} \ \left \| \mathbf{X}-\mathbf{D}\mathbf{Z} \right \|_F^2+2\lambda_2\sum_{c=1}^{C}f(\mathbf{Z},\boldsymbol{y}^c,\boldsymbol{u}_c,b_c)\\&+\lambda_1\left \| \mathbf{Z} \right \|_F^2, \  \textrm{s.t.} \ \left \| \boldsymbol{d}_k \right \|^2\leq 1
\end{split}
\end{equation}
where $\mathbf{U}=[\boldsymbol{u}_1,\boldsymbol{u}_2,\ldots,\boldsymbol{u}_C]$ and $\boldsymbol{b}=[b_1,b_2,\ldots,b_C]$.

\section{Proposed Method}
\label{sec:sec3}
In this section, our proposed LCDL-SV is presented. First we will introduce a locality constraint on the atoms of the learned dictionary. Then by incorporating the locality constraint and the support vector discriminative term into the framework of SDL, we will present the formulations of our proposed LCDL-SV.

\subsection{Locality constraint on atoms}
As mentioned earlier, $\mathbf{Z}$ is the coding coefficient matrix of training data $\mathbf{X}$ over the dictionary $\mathbf{D}$, and  $\boldsymbol{z}_i=[\boldsymbol{z}_{1,i},\boldsymbol{z}_{2,i},\ldots,\boldsymbol{z}_{K,i}]^T,(i=1,2,\ldots,n)$ is the coding vector of $\boldsymbol{x}_i$ on $\mathbf{D}$. The input training data can be represented as a linear combination of atoms in the dictionary, and the formulation is illustrated in Fig. \ref{fig:atom_profile}.
\Figure[h!](topskip=0pt, botskip=0pt, midskip=0pt)[width=3 in]{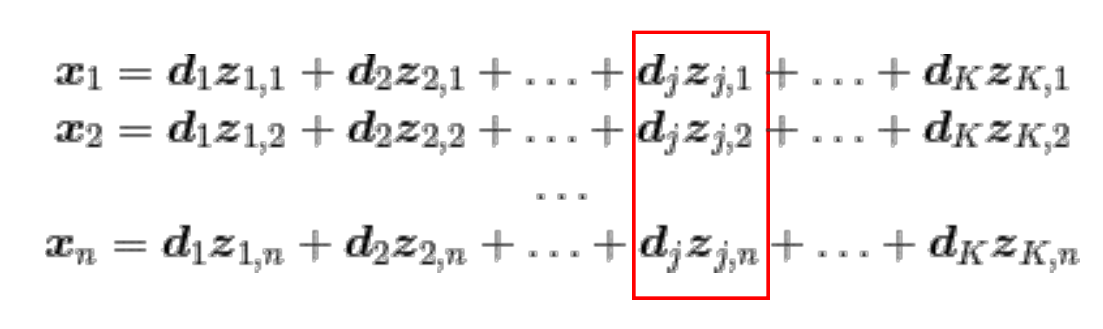}
{Linear representation of training data by atoms in the dictionary.\label{fig:atom_profile}}

In \cite{b48} and \cite{b19}, the $j$-th row vector of $\mathbf{Z}$ is coined the profile of atom $\boldsymbol{d}_j$. Thus, $\hat{\boldsymbol{z}}_j=[\boldsymbol{z}_{j,1},\boldsymbol{z}_{j,2},\ldots,\boldsymbol{z}_{j,n}]^T(j=1,2,\ldots,K)$ is the profile of atom $\boldsymbol{d}_j$, and the red rectangle in Fig. \ref{fig:atom_profile} depicts profile $\hat{\boldsymbol{z}}_j$. So the profile matrix can be constructed as $\mathbf{Z}^T=[\hat{\boldsymbol{z}}_1,\hat{\boldsymbol{z}}_2,\ldots,\hat{\boldsymbol{z}}_K]\in\mathbb{R}^{n\times K}$, which is the transpose matrix of $\mathbf{Z}$. Based on the definition of profile, the linear representations in Fig. \ref{fig:atom_profile} can be reformulated as follows,
\begin{equation}
\label{eq:equ_repre}
\mathbf{X}=\boldsymbol{d}_1(\hat{\boldsymbol{z}}_1)^T+\ldots+\boldsymbol{d}_j(\hat{\boldsymbol{z}}_j)^T+\ldots+\boldsymbol{d}_K(\hat{\boldsymbol{z}}_K)^T
\end{equation}
From (\ref{eq:equ_repre}), one can see that the profile $\hat{\boldsymbol{z}}_j$ and atom $\boldsymbol{d}_j$ have a one-to-one correspondence. In this paper, instead of preserving locality information of the original training data, we introduce a locality constraint on the atoms of the learned dictionary, which has proven to be more effective and robust\cite{b19}. A viable way to encourage similar atoms to have similar profiles is to minimize the following problem,
\begin{equation}
\label{eq:locality_term}
\frac{1}{2}\sum_{i=1}^{K}\sum_{j=1}^{K}(\hat{\boldsymbol{z}}_i-\hat{\boldsymbol{z}}_j)^2\mathbf{M}_{ij}
\end{equation}
where $\mathbf{M}$ is a similarity matrix which can be defined as,
\begin{equation}
\label{eq:sim_graph}
\mathbf{M}_{ij}=\left\{\begin{matrix}
\textrm{exp}(-\frac{\left \| \boldsymbol{d}_i-\boldsymbol{d}_j \right \|_2}{\delta}),  \ \textrm{if} \  \boldsymbol{d}_j \in \textrm{\textit{k}NN}(\boldsymbol{d}_i)\\ 
\\
0, \ \textrm{otherwise}
\end{matrix}\right.
\end{equation}
where $\textrm{\textit{k}NN}(\boldsymbol{d}_i)$ represents the $k$-nearest neighbors of atom $\boldsymbol{d}_i$ and $\delta$ is a parameter. After some deductions, we can obtain the following equivalent formulation of (\ref{eq:locality_term}),
\begin{equation}
\label{eq:locality_term_eq}
\frac{1}{2}\sum_{i=1}^{K}\sum_{j=1}^{K}(\hat{\boldsymbol{z}}_i-\hat{\boldsymbol{z}}_j)^2\mathbf{M}_{ij}=tr(\mathbf{Z}^T\mathbf{L}\mathbf{Z})
\end{equation}
where $\mathbf{L}=\mathbf{T}-\mathbf{M}$ is a graph Laplacian matrix, $\mathbf{T}=\textrm{diag}(t_1,\ldots,t_K)$ and $t_i=\sum_{j=1}^{K}\mathbf{M}_{ij}$. We can observe that the graph Laplacian matrix $\mathbf{L}$ is defined on the learned dictionary $\mathbf{D}$. As a result, the graph Laplacian matrix $\mathbf{L}$ is updated due to the fact that $\mathbf{D}$ is updated in the dictionary learning process. Therefore, the graph Laplacian matrix $\mathbf{L}$ can inherit the manifold structure of the training samples.

\subsection{LCDL-SV model}
Apart from the locality constraint on the atoms, to facilitate the subsequent classification stage, a support vector discriminative term is incorporated into our proposed method. The purpose of this term is to enforce the coefficients of different classes to be separated by a max-margin. Intuitively, when the coefficients are separated by a hyperplane, the large margin of different classes can promote the confidence of classification. Moreover, the parameters of SVM (\ie, $\mathbf{U}$ and $\boldsymbol{b}$) can be learned in our dictionary learning process. The support vector discriminative term has the same formulation in (\ref{eq:svm_term}). Thus, the objective function of our proposed LCDL-SV is formulated as follows,
\begin{equation}
\label{eq:lcsvgdl}
\begin{split}
&\underset{\mathbf{D},\mathbf{Z},\mathbf{L},\mathbf{U},\boldsymbol{b}}{\textrm{min}} \ \left \| \mathbf{X}-\mathbf{D}\mathbf{Z} \right \|_F^2+2\lambda_2\sum_{c=1}^{C}f(\mathbf{Z},\boldsymbol{y}^c,\boldsymbol{u}_c,b_c)\\&+\lambda_1tr(\mathbf{Z}^T\mathbf{L}\mathbf{Z}), \  \textrm{s.t.} \ \left \| \boldsymbol{d}_k \right \|^2=1,k=1,\ldots,K
\end{split}
\end{equation}
where $\lambda_1$ and $\lambda_2$ are two balancing parameters.

\section{Optimization}
\label{sec:sec4}
In this section, we adopt an alternative strategy to solve the LCDL-SV model. The alternative minimization scheme can be partitioned into the following three sub-problems,

\textit{Update} $\mathbf{Z}$: Fix the other variables and update $\mathbf{Z}$ by solving the following problem:
\begin{equation}
\label{eq:Z_problem}
\begin{split}
&\underset{\mathbf{Z}}{\textrm{min}} \ \left \| \mathbf{X}-\mathbf{D}\mathbf{Z} \right \|_F^2+2\lambda_2\theta\sum_{c=1}^{C}l(\boldsymbol{z}_i,\boldsymbol{y}_i^{c},\boldsymbol{u}_c,b_c)\\&+\lambda_1tr(\mathbf{Z}^T\mathbf{L}\mathbf{Z})
\end{split}
\end{equation}
The optimization of $\mathbf{Z}$ in (\ref{eq:Z_problem}) can be performed by columns, which is formulated as,
\begin{equation}
\label{eq:problem_z}
\begin{split}
&\underset{\boldsymbol{z}_i}{\textrm{min}} \ \left \| \boldsymbol{x}_i-\mathbf{D}\boldsymbol{z}_i \right \|_2^2+2\lambda_2\theta\sum_{c=1}^{C}l(\boldsymbol{z}_i,y_i^c,\boldsymbol{u}_c,b_c)\\&+\lambda_1tr(\boldsymbol{z}_i^T\mathbf{L}\boldsymbol{z}_i)
\end{split}
\end{equation}
To facilitate the optimization process, we employ the quadratic hinge loss function to approximate the original one. The quadratic hinge loss function is defined as,
\begin{equation}
\label{eq:hingeloss}
\begin{split}
&l(\boldsymbol{z}_i,y_i^c,\boldsymbol{u}_c,b_c)=\\&\left\{\begin{matrix}
\left \| y_i^c(\boldsymbol{u}_c^T\boldsymbol{z}_i+b_c)-1 \right \|_2^2, \ y_i^c(\boldsymbol{u}_c^T\boldsymbol{z}_i+b_c)-1> 0\\
0,t=1 \ \textrm{or} \ y_i^c(\boldsymbol{u}_c^T\boldsymbol{z}_i+b_c)-1\leq 0 
\end{matrix}\right.
\end{split}
\end{equation}
where $t$ denotes the iteration number. When $t$=1, (\ref{eq:problem_z}) is degenerated into the following problem, 
\begin{equation}
\label{eq:problem_t1}
\underset{\boldsymbol{z}_i}{\textrm{min}} \ \left \| \boldsymbol{x}_i-\mathbf{D}\boldsymbol{z}_i \right \|_2^2+\lambda_1tr(\boldsymbol{z}_i^T\mathbf{L}\boldsymbol{z}_i)
\end{equation}
(\ref{eq:problem_t1}) has the following closed-form solution,
\begin{equation}
\label{eq:solution_t1}
\boldsymbol{z}_i=(\mathbf{D}^T\mathbf{D}+\lambda_1\mathbf{L})^{-1}\mathbf{D}^T\boldsymbol{x}_i
\end{equation}
When $t\ge$2, (\ref{eq:problem_z}) can be rewritten as, 
\begin{equation}
\label{eq:problem_t2}
\begin{split}
&\underset{\boldsymbol{z}_i}{\textrm{min}} \ \left \| \boldsymbol{x}_i-\mathbf{D}\boldsymbol{z}_i \right \|_2^2+\lambda_1tr(\boldsymbol{z}_i^T\mathbf{L}\boldsymbol{z}_i)\\&+2\lambda_2\theta\sum_{c\in \phi}\left \| y_i^c(\boldsymbol{u}_c^T\boldsymbol{z}_i+b_c)-1 \right \|_2^2
\end{split}
\end{equation}
where $\phi=\left \{ c|1\leq c\leq C,y_i^c(\boldsymbol{u}_c^T\boldsymbol{z}_i+b_c)-1>0 \right \}$, (\ref{eq:problem_t2}) also has closed-form solution which is given by,
\begin{equation}
\label{eq:solution_t2}
\boldsymbol{z}_i=(\mathbf{D}_1)^{-1}\mathbf{D}_2
\end{equation}
where $\mathbf{D}_1=\mathbf{D}^T\mathbf{D}+\lambda_1\mathbf{L}+2\lambda_2\theta \sum_{c\in \phi}u_cu_c^T$ and $\mathbf{D}_2=\mathbf{D}^T\boldsymbol{x}_i+2\lambda_2\theta\sum_{c\in \phi}\boldsymbol{u}_c(y_i^c-b_c)$

\textit{Update} $\mathbf{D}$ and $\mathbf{L}$: To update $\mathbf{D}$, we fix variables other than $\mathbf{D}$ and minimize (\ref{eq:lcsvgdl}), which leads to
\begin{equation}
\label{eq:problem_d}
\underset{\mathbf{D}}{\textrm{min}} \ \left \| \mathbf{X}-\mathbf{D}\mathbf{Z} \right \|_F^2, \textrm{s.t.} \ \left \| \boldsymbol{d}_k \right \|^2=1,k=1,\ldots,K
\end{equation}
We can see that (\ref{eq:problem_d}) becomes a least squares problem with quadratic constraints. Here we employ the Lagrange dual function \cite{b49} to solve (\ref{eq:problem_d}), and the Lagrange dual function of (\ref{eq:problem_d}) is formulated as,
\begin{equation}
\label{eq:lagrange}
g(\boldsymbol{\delta})=\underset{\mathbf{D}}{\textrm{inf}}\left ( \left \| \mathbf{X}-\mathbf{D}\mathbf{Z} \right \|_F^2+ \sum_{k=1}^{K}\delta_k(\left \| \boldsymbol{d}_k \right \|^2-1)\right )
\end{equation}
where $\boldsymbol{\delta}=[\delta_1,\delta_2,\ldots,\delta_K]$ and $\delta_k$ is the Lagrange multiplier corresponds to the $k$-th equality constraint ($\left \| \boldsymbol{d}_k \right \|^2-1=0$). We can define a diagonal matrix $\mathbf{\Delta}$ whose diagonal element $\mathbf{\Delta}_{kk}=\delta_k$, then (\ref{eq:lagrange}) can be reformulated as,
\begin{equation}
\label{eq:equi}
L(\mathbf{D},\boldsymbol{\delta})=\left \| \mathbf{X}-\mathbf{D}\mathbf{Z} \right \|_F^2+tr(\mathbf{D}^T\mathbf{D}\mathbf{\Delta})-tr(\mathbf{\Delta})
\end{equation}
By setting the first-order derivative of (\ref{eq:equi}) to zero, we can obtain the following solution to $\mathbf{D}$,
\begin{equation}
\label{eq:solution_d}
\mathbf{D}=\mathbf{X}\mathbf{Z}^T(\mathbf{Z}\mathbf{Z}^T+\mathbf{\Delta})^{-1}
\end{equation}
To speed up the optimization process, we discard $\mathbf{\Delta}$ in the final formulation, which is given by,
\begin{equation}
\label{eq:solution_d_f}
\mathbf{D}=\mathbf{X}\mathbf{Z}^T(\mathbf{Z}\mathbf{Z}^T)^{-1}
\end{equation}
When $\mathbf{D}$ is updated, we update the graph Laplacian matrix $\mathbf{L}$ by using (\ref{eq:sim_graph}).

\textit{Update} $\mathbf{U}$ and $\boldsymbol{b}$: When the other variables are fixed, (\ref{eq:lcsvgdl}) with respect to $\mathbf{U}$ and $\boldsymbol{b}$ is boiled down to the following problem,
\begin{equation}
\label{eq:svm_problem}
\underset{\mathbf{U},\boldsymbol{b}}{\textrm{min}} \ \sum_{c=1}^{C}\left \{ \left \| \boldsymbol{u}_c \right \|_2^2+\theta\sum_{i=1}^{n}l(\boldsymbol{z}_i,y_i^c,\boldsymbol{u}_c,b_c) \right \}
\end{equation}
(\ref{eq:svm_problem}) is a multi-class linear SVM problem which can be solved by the SVM solver presented in\cite{b50}.
Due to the fact that the objective function proposed in (\ref{eq:lcsvgdl}) is non-convex, the algorithm cannot converge to the global minimum. However, satisfactory solutions can be obtained with the decreasing of the objective function. The convergence curve of LCDL-SV on the Extended Yale B database is plotted in Fig. \ref{fig: converg_curve}. Algorithm \ref{alg1} outlines the optimization process of our proposed LCDL-SV.
\Figure[h!](topskip=0pt, botskip=0pt, midskip=0pt)[width=3 in]{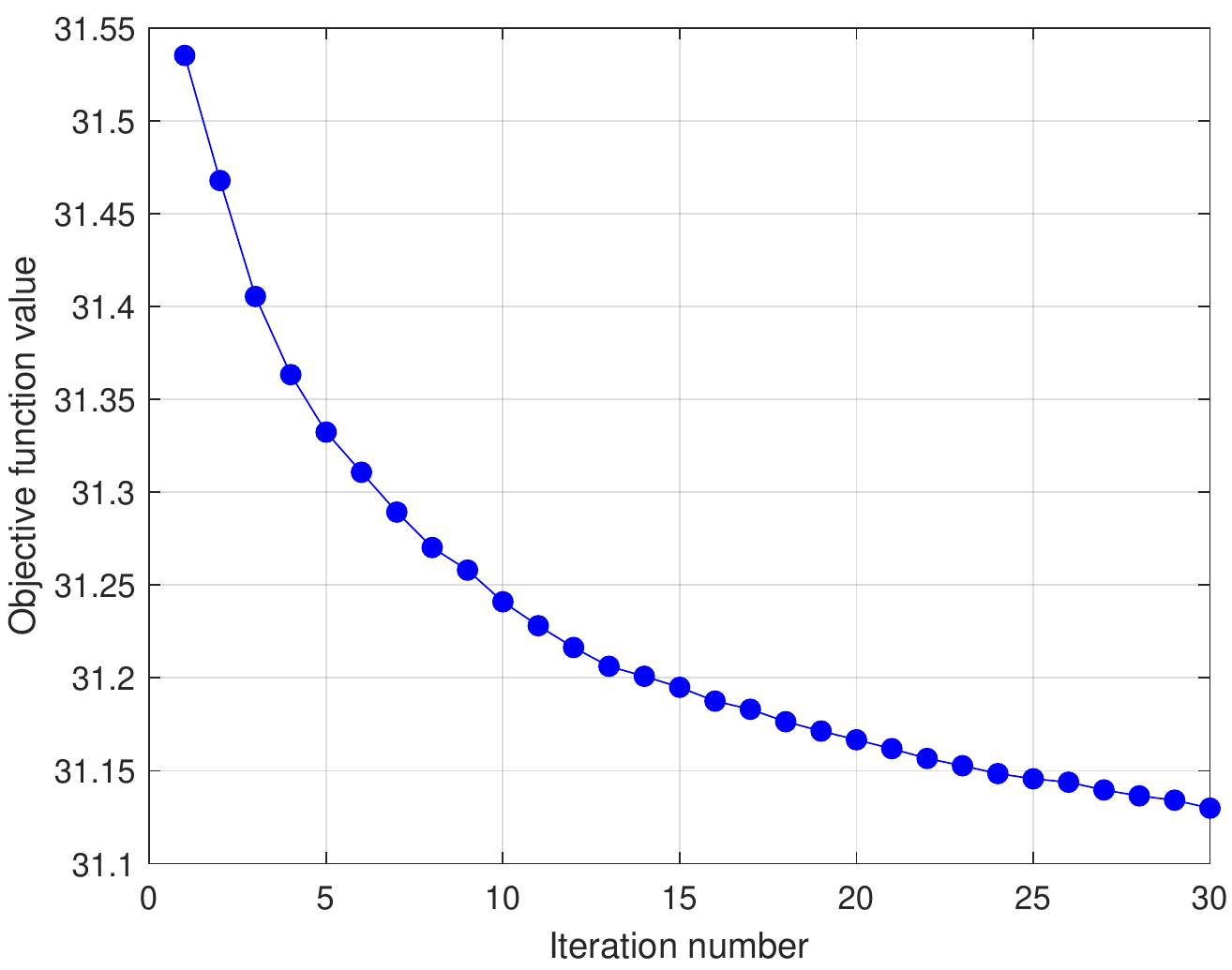}
{Convergence curve of LCDL-SV on the Extended Yale B database.\label{fig: converg_curve}}

\begin{algorithm} 
\caption{Optimization procedure of LCDL-SV} 
\label{alg1} 
\begin{algorithmic}[1]
\REQUIRE Training data matrix $\mathbf{X}$, parameters $\lambda_1$, $\lambda_2$ and $\theta$
\STATE Initialize $\mathbf{D}$, $\mathbf{Z}$, $\mathbf{U}$ and $\boldsymbol{b}$
\WHILE{not converged} 
\STATE Construct the graph Laplacian matrix $\mathbf{L}$ by using (\ref{eq:sim_graph})
\FOR{$i$=1 to $n$}  
\STATE Update $\boldsymbol{z}_i$ by using (\ref{eq:solution_t1}) and (\ref{eq:solution_t2})
\ENDFOR
\STATE Update the dictionary by using (\ref{eq:solution_d_f})
\FOR{$c$=1 to $C$}  
\STATE Update $\mathbf{U}_c$ and $b_c$ by solving (\ref{eq:svm_problem})
\ENDFOR
\ENDWHILE 
\ENSURE $\mathbf{D}$, $\mathbf{U}$ and $\boldsymbol{b}$ 
\end{algorithmic} 
\end{algorithm}
When the dictionary learning process is completed, we perform classification as follows. For a test sample $\boldsymbol{x}_{new}$, first we obtain its coding vector by $\boldsymbol{z}=\mathbf{P}\boldsymbol{x}_{new}$, where $\mathbf{P}=(\mathbf{D}^T\mathbf{D}+\eta_1\mathbf{I})^{-1}\mathbf{D}^T$. Then the regularized residual for the $c$-th class can be obtained by,
\begin{equation}
\label{eq:c_residual}
\boldsymbol{r}_c=\frac{\left \| \boldsymbol{x}_{new}-\mathbf{D}_c\boldsymbol{z}_c \right \|_2}{\left \| \boldsymbol{z}_c \right \|_2}
\end{equation}
where $\mathbf{D}_c$ and $\boldsymbol{z}_c $ are the sub-dictionary and coding vector associated with the $c$-th class, respectively. Moreover, the result produced by the learned SVM classifier is formulated as,
\begin{equation}
\label{eq:c_svm}
\boldsymbol{s}_c=\boldsymbol{u}_c^T\boldsymbol{z}+b_c
\end{equation}
Finally, the identity of $\boldsymbol{x}_{new}$ is given by,
\begin{equation}
\label{eq:class_rule}
\textrm{identity}(\boldsymbol{x}_{new})=\underset{c}{\textrm{min}} \ (\boldsymbol{r}_c-\eta_2\boldsymbol{s}_c)
\end{equation}
where $\eta_2$ is a weighting parameter.

\section{Experimental results}
\label{sec:sec5}
In this section, we conduct experiments on five publicly available databases, \ie, the Extended Yale B database\cite{b51}, AR database\cite{b52}, Scene 15 dataset\cite{b53}, Caltech 101 dataset\cite{b54} and LFW database\cite{b55}. We compare LCDL-SV with SRC\cite{b56}, D-KSVD\cite{b12}, LC-KSVD\cite{b13}, FDDL\cite{b6}, SVGDL\cite{b14} and two recently proposed ADL approaches, \ie, CADL\cite{b38} and SADL\cite{b36}. To validate the effectiveness of employing both the regularized residual and SVM, we also report the results of LCDL-SV only using regularized residual for classification and LCDL-SV only employing the learned multi-class SVM for classification, which are denoted by LCDL-SV (Res) and LCDL-SV (SVM), respectively. Besides the classification accuracy, we also record the training time and testing time of these competing methods in our experiments. SRC directly employs all the training data as the dictionary, thus we do not report its training time. The difference between LCDL-SV (Res), LCDL-SV (SVM) and LCDL-SV lies in the classification rule, thus they have the same training time but different testing time. All experiments are conducted with MATLAB R2019b under Windows 10 on a PC equipped with Intel i9-8950HK 2.90 GHz CPU and 32 GB RAM.

There are five parameters in our proposed method, \ie, $\theta$, $\lambda_1$, $\lambda_2$, $\eta_1$ and $\eta_2$. In all experiments, $\theta$ is set to be 0.2, the other four parameters are determined by cross-validation, and $\lambda_1$ and $\lambda_2$ are selected from $10^{-6},10^{-5},\ldots,10^{-1}$. The optimal values on each dataset are recorded in Table \ref{table:param_value}. For fair comparison, we tune the parameters of competing approaches to achieve their best performance.
\begin{table}[]
\centering
\caption{Optimal parameters of LCDL-SV on each dataset.}
\label{table:param_value}
\begin{tabular}{p{22pt} p{22pt} p{22pt} p{22pt} p{22pt} p{22pt} p{25pt}}
\hline
             & EYaleB & AR & Scene 15 & Caltech 101 & LFW (VGG16) & LFW (VGG19) \\ \hline
$\lambda_1$            &  1e-3 & 1e-3 &   1e-5  & 1e-1   &  1e-2  &   1e-2 \\
$\lambda_2$         &  1e-6 &  1e-6  &1e-4    & 1e-6   & 1e-6   &   1e-6 \\
$\eta_1$        & 1e-2  &  1e-3  &  1e-5  & 1   &  1e-2  &  1e-2  \\
$\eta_2$       &  5 & 50   &  10  & 600   &  600  &  80 \\ \hline
\end{tabular}
\end{table}

\subsection{Extended Yale B}
\label{sec:sect51}
The Extended Yale B database contains 2414 frontal face images of 38 subjects, each person has about 64 images, and some example images are shown in Fig. \ref{fig:exam_YaleB}. Following the experimental setting in \cite{b14}, in our experiments, all images are cropped to 54$\times$48, then they are reduced to a dimension of 300 by PCA. We randomly select 20 images per person as training set and the remaining as testing set. The dictionary has 380 atoms, which corresponds to an average of 10 atoms per subject. Experimental results are summarized in Table \ref{table:EYaleB}. We can observe that the proposed LCDL-SV achieves the highest recognition accuracy. Moreover, LCDL-SV is much faster than FDDL in the training phase, and the training time of LCDL-SV is comparable to that of SVGDL. Thanks to the framework of ADL, CADL and SADL are efficient on this dataset. Due to the jointly learning dictionary and multi-class SVM, the testing time of SVGDL and LCDL-SV (SVM) is less than our proposed LCDL-SV. Nevertheless, by fusing the regularized residual and the learned multi-class SVM, LCDL-SV outperforms all the competing approaches in recognition accuracy, and it is more efficient than SRC and FDDL in terms of testing time. It should be noted that, in \cite{b36}, the reported accuracy of SADL and SRC is 96.35\% and 96.51\%, respectively. The differences lie in the following two aspects. On the one hand, in \cite{b36}, each image of 192$\times$168 pixels is projected onto a 504-dimensional vector by random projection, while we use the cropped image of 54$\times$48 pixels and employ PCA to reduce the image to a dimension of 300. On the other hand, half of the images (i.e., 32 images) per subject are used for training in \cite{b36}, while 20 images per person are employed for training in our experiments.
\Figure[h!](topskip=0pt, botskip=0pt, midskip=0pt)[width=3 in]{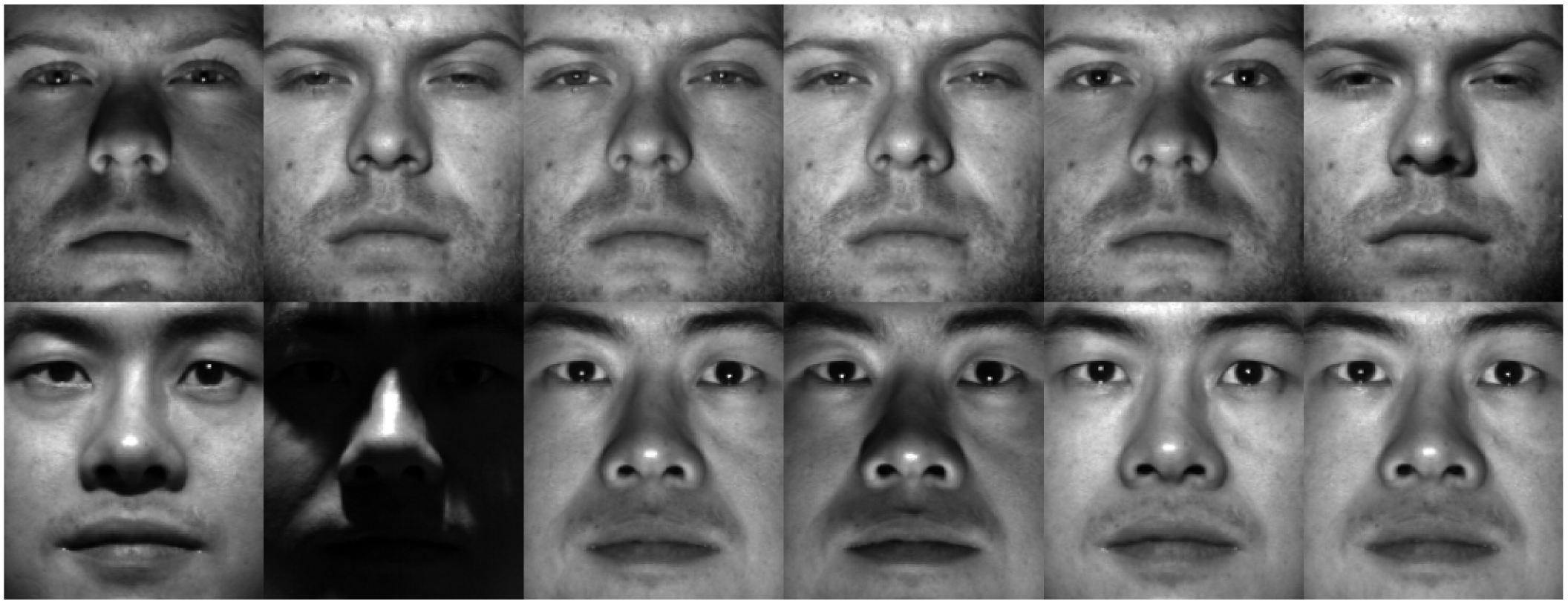}
{Example images from the Extended Yale B database.\label{fig:exam_YaleB}}

\begin{table}[]
\centering
\caption{Recognition accuracy (\%) and computing time on the Extended Yale B database.}
\label{table:EYaleB}
\begin{tabular}{cccc}
\hline
Methods & Accuracy & Training time(s)  & Testing time(s)\\ \hline
SRC\cite{b56}    & 90.0   &  No Need &  3.2 \\
D-KSVD\cite{b12}    &  75.3 & 7.8  &   0.2   \\
LC-KSVD\cite{b13}    &  90.6 & 11.5  &  0.2    \\
FDDL\cite{b6}     &  91.9  & 592.6  & 3.4     \\
SVGDL\cite{b14}     &  96.1 &  57.8 &   0.01   \\
CADL\cite{b38}     &   96.3 & 23.8  &  0.01   \\
SADL\cite{b36}     &   95.2 & 27.2  &  0.01   \\
LCDL-SV (Res)     &  91.5 & 57.2  &   0.8   \\
LCDL-SV (SVM)     &   97.1 & 57.2  &   0.06  \\
LCDL-SV    &  \textbf{97.8} & 57.2  &  0.9   \\ \hline
\end{tabular}
\end{table}

\subsection{AR}
The AR database has more than 4000 face images of 126 subjects with variations in facial expression, illumination conditions and occlusions. Fig. \ref{fig:exam_AR} shows example images from the database. In our experiments, we use a subset of 2600 images of 50 male and 50 female subjects from the database. As in \cite{b13}, each 165$\times$120 face image is projected onto a 540-dimensional vector by random projection. For each person, 20 images are randomly selected for training and the remaining for testing. The learned dictionary has 500 atoms, namely five atoms per person. Table \ref{table:AR} lists the recognition accuracy and computing time of all compared methods. Notice that 20 atoms per class are exploited in \cite{b38}, while only five atoms per class are used in our experiments. One can see that LCDL-SV has the best performance in recognition accuracy and is more efficient than FDDL in both training and testing phases. Moreover, on this database and the Extended Yale B database, LCDL-SV (SVM) achieves better results than SVGDL, which demonstrates that locality constraint of atoms does promote the classification performance of SDL approaches on these two face databases.
\Figure[h!](topskip=0pt, botskip=0pt, midskip=0pt)[width=3 in]{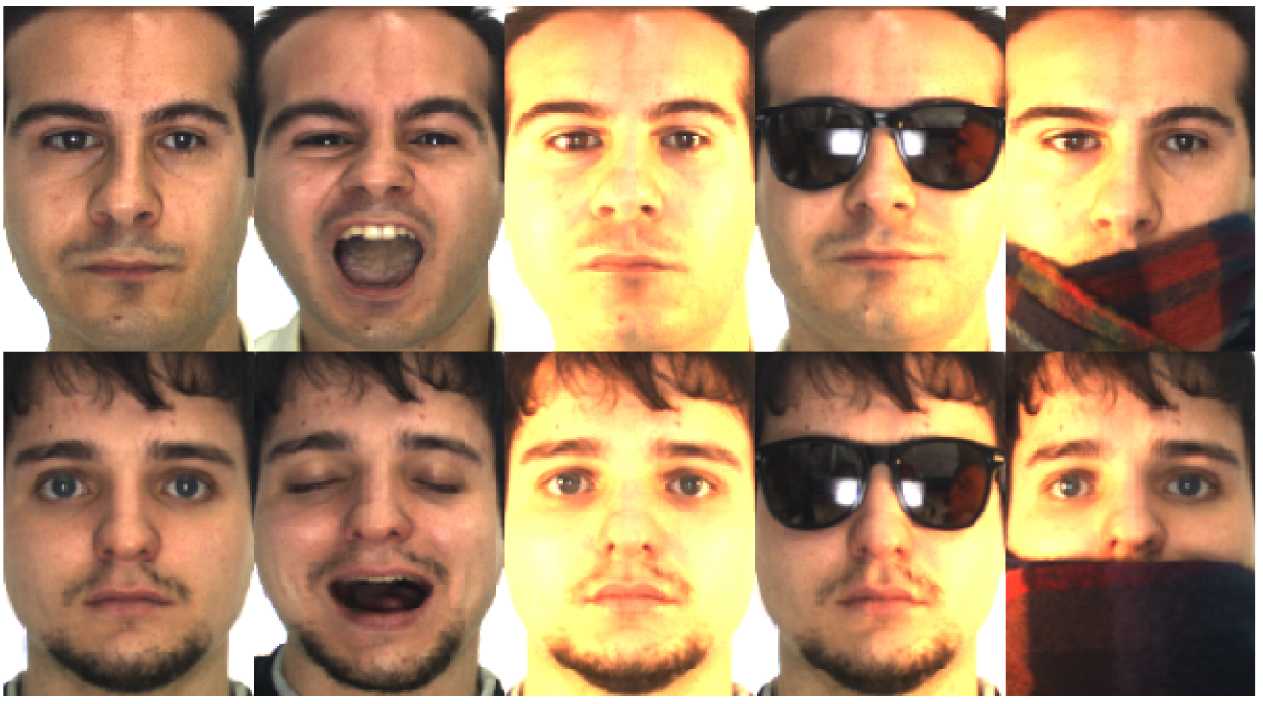}
{Example images from the AR database.\label{fig:exam_AR}}

\begin{table}[]
\centering
\caption{Recognition accuracy (\%) and computing time on the AR database.}
\label{table:AR}
\begin{tabular}{cccc}
\hline
Methods & Accuracy & Training time(s) & Testing time(s)\\ \hline
SRC\cite{b56}    &  66.5  & No Need & 2.3  \\
D-KSVD\cite{b12}    &  88.8  & 25.3 &  0.08  \\
LC-KSVD\cite{b13}    &    93.7 & 38.2 & 0.1  \\
FDDL\cite{b6}     &  97.4 &  822.4 &  2.9   \\
SVGDL\cite{b14}     &  98.8 & 358.9 &  0.01   \\
CADL\cite{b38}     &   98.5 & 288.3 &   0.01 \\
SADL\cite{b36}     &   97.2 & 183.1  & 0.01   \\
LCDL-SV (Res)     &  94.7 & 360.3 &  0.6   \\
LCDL-SV (SVM)     &   99.0 & 360.3 &  0.01  \\
LCDL-SV     &  \textbf{99.2} & 360.3 &  0.62   \\ \hline
\end{tabular}
\end{table}

\subsection{Scene 15}
Scene 15 dataset contains 15 natural scene categories, which comprises a wide range of indoor and outdoor scenes, such as bedroom, office and mountain, example images from this dataset are shown in Figure~\ref{fig:exam_scene15}. For fair comparison, we employ the 3000-dimensional SIFT-based features used in LC-KSVD\cite{b13}. Following the common experimental settings, we randomly select 100 images per category as training data and use the remaining for testing. The learned dictionary has 450 atoms. Experimental results are shown in Table \ref{table:scene15}. The recognition accuracy of LCDL-SV is 99.0\%, which outperforms all the compared methods. CADL performs the second best on this dataset, followed by SADL and SVGDL. Moreover, LCDL-SV is 15 times faster than FDDL in the training stage. The confusion matrix for LCDL-SV is depicted in Fig. \ref{fig:s15_confusion}, in which diagonal elements are well-marked. It can be seen that LCDL-SV attains 100\% recognition accuracy for the categories of suburb, forest and inside-city.
\Figure[h!](topskip=0pt, botskip=0pt, midskip=0pt)[width=3 in]{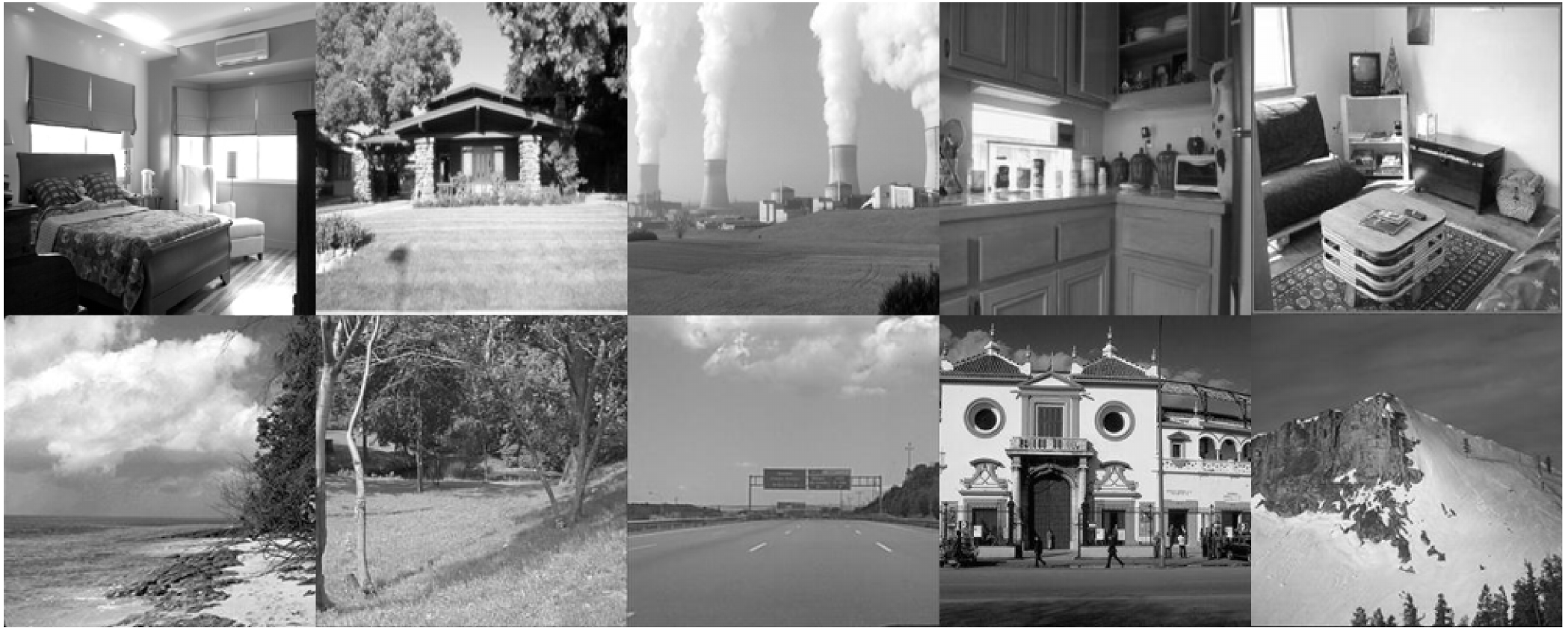}
{Example images from the Scene 15 dataset.\label{fig:exam_scene15}}

\begin{table}[]
\centering
\caption{Recognition accuracy (\%) and computing time on the Scene 15 dataset.}
\label{table:scene15}
\begin{tabular}{cccc}
\hline
Methods & Accuracy & Training time(s)  & Testing time(s)\\ \hline
SRC\cite{b56}    &   91.8 &  No Need & 3.5  \\
D-KSVD\cite{b12}    &   89.1  &  54.1 & 0.4   \\
LC-KSVD\cite{b13}    &   92.9  & 72.2  & 0.4   \\
FDDL\cite{b6}     &  97.5 & 2470.5  &  208.2    \\
SVGDL\cite{b14}     &   98.4 & 159.0  &  0.1   \\
CADL\cite{b38}     &   98.6 & 5365.8  & 0.67    \\
SADL\cite{b36}     &   98.5 & 314.7  &  0.2   \\
LCDL-SV (Res)     &  97.8 & 159.2  &  6.5    \\
LCDL-SV (SVM)     &    98.4  & 159.2  & 0.4  \\
LCDL-SV     &  \textbf{99.0} & 159.2  &  6.7    \\ \hline
\end{tabular}
\end{table}

\Figure[h!](topskip=0pt, botskip=0pt, midskip=0pt)[width=3 in]{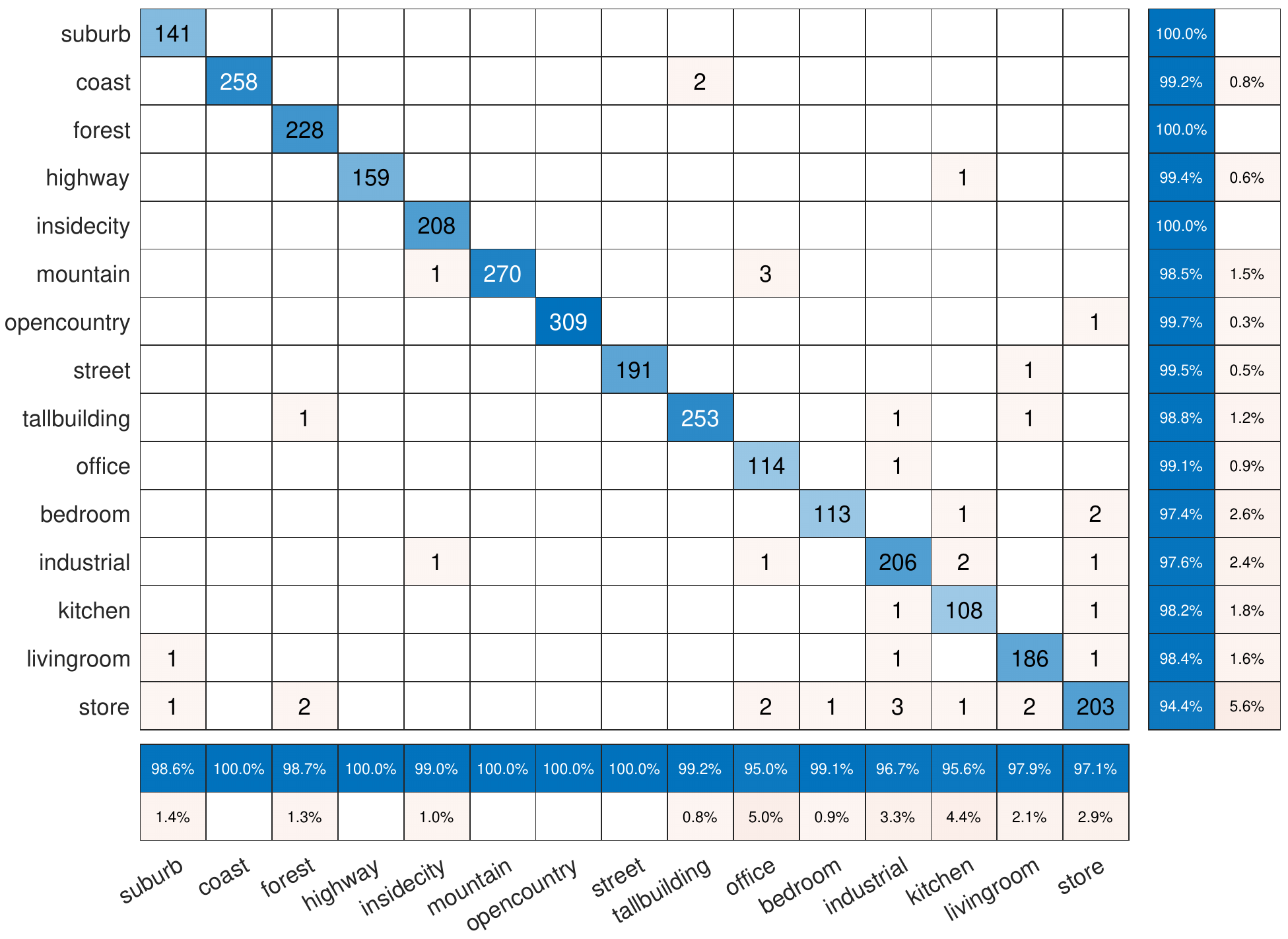}
{Confusion matrix on the Scene 15 dataset.\label{fig:s15_confusion}}

\subsection{Caltech 101}
Caltech101 dataset is a widely used dataset for object classification with 102 classes (i.e., 101 object classes and one background class). The number of images in each category is unbalanced, varying from 31 to 800, and in total this dataset contains 9144 images. For fair comparison, we also employ the 3000-dimensional SIFT-based features used in LC-KSVD\cite{b13}. Following the common experimental protocol, we randomly choose 5, 10, 15, 20, 25, and 30 samples per category for training and test on the remaining images. We repeat this process 10 times with different splits of training and test images and record the averaged classification accuracy. Table \ref{table:caltech} summarizes the classification results and Table \ref{table:time_caltech} lists the training time and testing time when 5 samples per category are used for training. As can be seen from Table \ref{table:caltech}, LCDL-SV consistently outperforms the other competing approaches in all cases. Compared with SVGDL, LCDL-SV (SVM) does not always achieve better accuracy (\eg, when the number of training samples per category is 25). This indicates that using only the locality constraint cannot guarantee the best performance. Combining with the proposed classification scheme, LCDL-SV exhibits its advantage over other dictionary learning approaches. From Table \ref{table:time_caltech}, we can observe that the training time of LCDL-SV is only one-sixteenth of that of FDDL and LCDL-SV is faster than SRC in the testing phase.
\Figure[h!](topskip=0pt, botskip=0pt, midskip=0pt)[width=3 in]{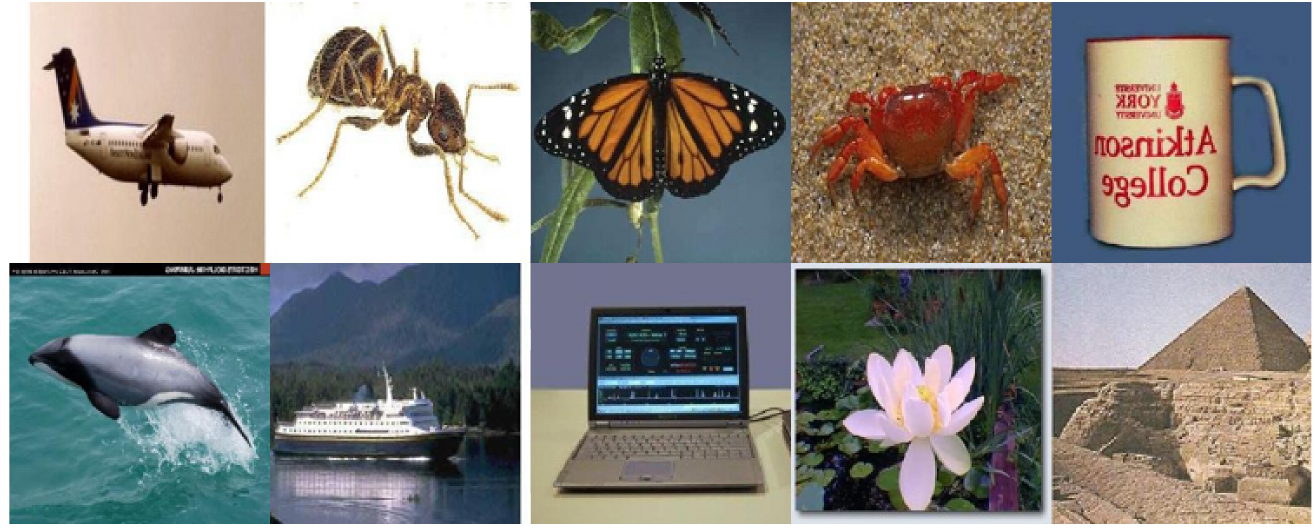}
{Example images from the Caltech 101 dataset.\label{fig:exam_cal101}}

\begin{table}[]
\centering
\caption{Recognition accuracy (\%) on the Caltech 101 dataset.}
\label{table:caltech}
\begin{tabular}{ccccccc}
\hline
number of train. samp.   & 5 & 10 & 15 & 20 & 25 & 30 \\ \hline
SRC\cite{b56}            &  48.8 & 60.1  & 64.9    &  67.7  & 69.2   &  70.7  \\
D-KSVD\cite{b12}         &  49.6 &  59.5  &  65.1  &  68.6  &  71.1  &  73.0  \\
LC-KSVD\cite{b13}        &  54.0 &  63.1  & 67.7   &  70.5  & 72.3   &  73.6  \\
FDDL\cite{b6}           & 53.6  &  63.6  & 66.8   & 69.8   &  71.7  & 73.1   \\
SVGDL\cite{b14}          & 55.3  &  64.3  &  69.6  &  72.3  &  \textbf{75.1}  & 76.7 \\
CADL\cite{b38}          & 55.6  &  63.9  &  65.7  &  68.1  &  70.1  & 75.0 \\
SADL\cite{b36}          & 46.6  &  59.4  & 62.4   &  68.4  &  70.3  & 74.4 \\
LCDL-SV (Res) & 52.6  & 61.1   &  62.2  &  63.9  & 62.2   &  60.7  \\
LCDL-SV (SVM) & 56.9  & 65.6   & 69.2   & 72.6   &  74.5  & 76.5   \\
LCDL-SV       & \textbf{57.0}  &  \textbf{65.9}  &  \textbf{69.7}  &  \textbf{73.1}  &  \textbf{75.1}  &  \textbf{76.8}  \\ \hline
\end{tabular}
\end{table}

\begin{table}[]
\centering
\caption{Computing time on the Caltech 101 dataset when 5 samples per category are used for training.}
\label{table:time_caltech}
\begin{tabular}{ccc}
\hline
Methods & Training time(s)  & Testing time(s)\\ \hline
SRC\cite{b56}     &  No Need &  33.9 \\
D-KSVD\cite{b12}     & 35.3  & 1.3   \\
LC-KSVD\cite{b13}   & 43.3  &  1.6  \\
FDDL\cite{b6}    & 2112.7  &  177.2    \\
SVGDL\cite{b14}   & 162.3  & 0.3    \\
CADL\cite{b38}  & 183.6  &   1.2  \\
SADL\cite{b36}   & 176.2  & 0.2    \\
LCDL-SV (Res)    & 139.4  &  23.2    \\
LCDL-SV (SVM)     & 139.4  & 2.1  \\
LCDL-SV   &  139.4    &   24.2  \\ \hline
\end{tabular}
\end{table}

\subsection{Deep features}
In this subsection, a subset of LFW database is used to evaluate our proposed LCDL-SV and other competing approaches on deep features. This subset contains 1251 images of 86 subjects, each person has 11-20 images. All images are converted to grayscale images and cropped and resized to 32$\times$32, some example images are shown in Fig. \ref{fig:exam_lfw}. Five images per subject are randomly selected as training samples and the remaining as test samples. The pre-trained VGG16 and VGG19 \cite{b57} models are employed to extract deep features, and FC6 in both VGG16 and VGG19 is used for feature extraction. The dimension of deep features extracted by VGG16 and VGG19 is 4096. In order to obtain more compact representations, we apply principal component analysis (PCA) to the 4096-dimensional features (keeping 98\% of the variance) and the dimensions of reduced features for VGG16 and VGG19 are 270 and 264, respectively. Finally, the reduced features are fed into the compared approaches. Experimental results are shown in Table \ref{table:lfw}, and the training time and testing time are recorded for the VGG19 feature. From Table \ref{table:lfw}, we can see that LCDL-SV delivers the best result on both the VGG16 and VGG19 features. This demonstrates that LCDL-SV is not only superior to its competing methods on hand-crafted features, but on the deep features as well. Similar to the observations on the other datasets used in our experiments, LCDL-SV is more efficient than FDDL in terms of training and testing time.
\Figure[h!](topskip=0pt, botskip=0pt, midskip=0pt)[width=3 in]{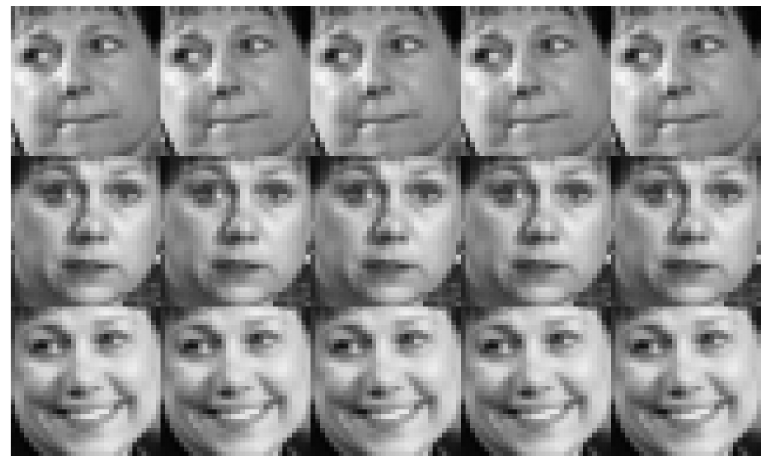}
{Example images from the LFW database.\label{fig:exam_lfw}}

\begin{table}[!h]
\centering
\caption{Recognition accuracy (\%) and computing time on the LFW database.}
\label{table:lfw}
\begin{tabular}{ccccc}
\hline
Methods  & VGG16 & VGG19 & Training time(s) & Testing time(s) \\ \hline
SRC\cite{b56}            &  42.4     &   36.4   & No Need  &  2.3  \\
D-KSVD\cite{b12}         &   35.7    &   35.8    & 5.9  & 0.2   \\
LC-KSVD\cite{b13}        &   41.5    &    36.9   & 8.7  &  0.1  \\
FDDL\cite{b6}           &   45.4    &     39.4   &  203.7 &  2.3 \\
SVGDL\cite{b14}          &   46.9    &     42.8  & 85.5  & 0.01   \\
CADL\cite{b38}          &  46.5     &    41.2   &  26.8 & 0.01   \\
SADL\cite{b36}          &   41.0    &   35.1   & 7.1  &  0.01   \\
LCDL-SV (Res) &  25.6     &    22.9  &  66.0 & 0.5    \\
LCDL-SV (SVM) &  47.6     &    45.3  &  66.0 &  0.03   \\
LCDL-SV       &  \textbf{48.0}     &    \textbf{46.5}  & 66.0  &    0.6 \\ \hline
\end{tabular}
\end{table}

\subsection{Parameter Sensitivity analysis}
As mentioned earlier, five parameters should be determined in our proposed method, \ie, $\theta$, $\lambda_1$, $\lambda_2$, $\eta_1$ and $\eta_2$. For the two parameters $\lambda_1$ and  $\lambda_2$, we find that relatively small values (e.g., 1e-5) can guarantee our proposed method to achieve satisfactory results for pattern classification tasks. $\eta_1$ is used to obtain the coding coefficients of test samples and it is usually set to 1e-3. For diverse datasets, $\eta_2$ has a relatively wide range in fusing the regularized residual and the results of multi-class SVM. Therefore, the above observations can be treated as a rule of thumb for selecting parameters of the proposed LCDL-SV. To investigate the sensitivity of parameters, we carry out experiments on the Extended Yale B database, and the experimental settings are the same as that in Section \ref{sec:sect51}. When analyzing one parameter, we fix the other two parameters. Firstly, we fix the parameters $\lambda_2$ and $\eta_2$, and examine how the performance changes with varying $\lambda_1$. Fig. \ref{fig:params} (a) plots the recognition accuracy with varying $\lambda_1$. Similarly, Figs. \ref{fig:params} (b) and \ref{fig:params} (c) plot the results of varying $\lambda_2$ and $\eta_2$. As can be seen from Fig. \ref{fig:params} (a), when the value of $\lambda_1$ increases from $10^{-5}$ to $10^{-3}$, the accuracy of LCDL-SV is gradually increasing. However, the performance of LCDL-SV will degrade when the value of $\lambda_1$ is larger than 0.01. From Fig. \ref{fig:params} (b), we can see that LCDL-SV achieves stable performance when the value of $\lambda_2$ is in the range of [$10^{-7}$,$10^{-4}$]. With the increasing of $\lambda_2$, the performance drops to some extent. A larger $\lambda_2$ will reduce the discriminative ability of the support vector term, leading to degenerated performance. From Fig. \ref{fig:params} (c), one can see that accuracy of LCDL-SV has an increase with $\eta_2$ from 1 to 5, and then has a decline when $\eta_2$ continues increasing. On the Extended Yale B database, LCDL-SV has the best performance when $\eta_2$ is set to be 5. 
\Figure[h!](topskip=0pt, botskip=0pt, midskip=0pt)[width=3 in]{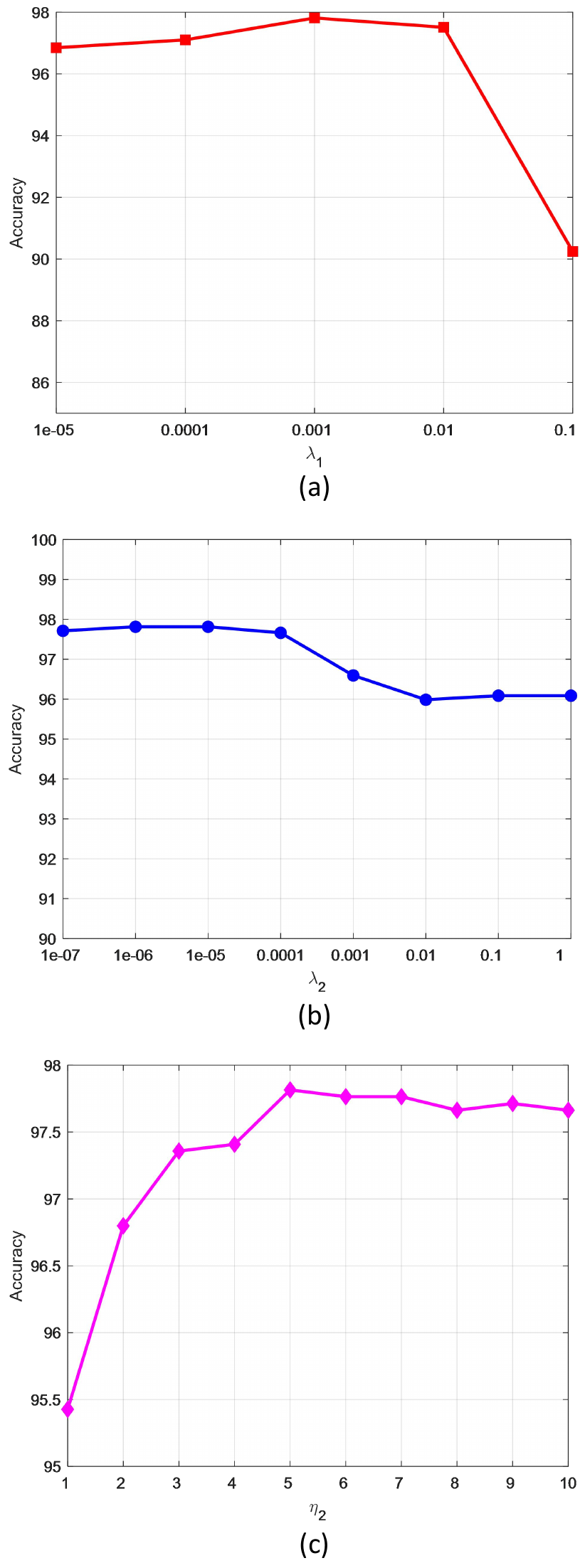}
{Classification accuracy (\%) of LCDL-SV with varying parameters on the Extended Yale B database. (a) $\lambda_1$ changes when $\lambda_2$ and $\eta_2$ are fixed to 1e-6 and 5, respectively. (b) $\lambda_2$ changes by fixing $\lambda_1$=1e-3 and $\eta_2$=5 and (c) $\eta_2$ varies when $\lambda_1$ and $\lambda_2$ are fixed to 1e-3 and 1e-6, respectively. \label{fig:params}}

\section{Conclusion}
\label{sec:sec6}
In this paper, we propose a locality constraint dictionary learning with support vector discriminative term (LCDL-SV) for pattern classification. In contrast with traditional methods in which the graph Laplacian matrix is derived from the original training data, we preserve the locality of atoms on the basis of the learned dictionary. By introducing a support vector discriminative term into the formulation of LCDL-SV, a classifier can be jointly learned in our dictionary learning procedures. More importantly, the regularized residual and multi-class SVM are simultaneously employed to classify the test samples. Experimental results on face databases, scene dataset and object dataset validate the effectiveness of LCDL-SV, and it outperforms some state-of-the-art dictionary learning approaches, \eg, FDDL, SVGDL, CADL, and SADL.

Discriminative analysis dictionary learning methods have aroused considerable interest due to their efficiency and efficacy. In future work, we will develop new discriminative ADL approach and apply it to other classification scenarios, such as action recognition and texture classification.

\begin{IEEEbiography}[{\includegraphics[width=1in,height=1.25in,clip,keepaspectratio]{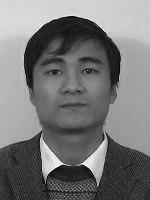}}]{HE-FENG YIN} received his B.S. degree in School of Computer Science and Technology from Xuchang University, Xuchang, China, in 2011. Currently, he is a PhD candidate in School of IoT Engineering, Jiangnan University, Wuxi, China. He was a visiting PhD student at centre for vision, speech and signal processing (CVSSP), University of Surrey, under the supervision of Prof. Josef Kittler. His research interests include representation based classification methods, dictionary learning and low rank representation.
\end{IEEEbiography}

\begin{IEEEbiography}[{\includegraphics[width=1in,height=1.25in,clip,keepaspectratio]{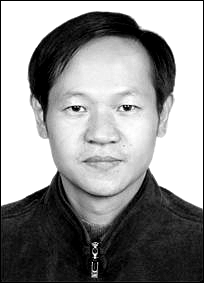}}]{XIAO-JUN WU} received the B.S. degree in mathematics from Nanjing Normal University, Nanjing, China, in 1991, and the M.S. and Ph.D. degrees in pattern recognition and intelligent system, from the Nanjing University of Science and Technology, Nanjing, in 1996 and 2002, respectively. From 1996 to 2006, he was with the School of Electronics and Information, Jiangsu University of Science and Technology, where he was involved in teaching and promoted to a Professor. From 1999 to 2000, he was a fellow of the International Institute for Software Technology, United Nations University. From 2003 to 2004, he was a Visiting Researcher with the Centre for Vision, Speech, and Signal Processing, University of Surrey, U.K. Since 2006, he has been with the School of Information Engineering, Jiangnan University, where he is currently a Professor of computer science and technology. He has published over 200 papers. His current research interests include pattern recognition, computer vision, and computational intelligence. He received the Most Outstanding Postgraduate Award from the Nanjing University of Science and Technology.
\end{IEEEbiography}

\begin{IEEEbiography}[{\includegraphics[width=1in,height=1.25in,clip,keepaspectratio]{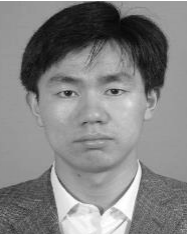}}]{Su-gen Chen} received the B.S. degree in mathematics and applied mathematics from Anqing Normal University, Anqing, Anhui, in 2004, the M.S. degree in computational mathematics from the Hefei University of Technology, Hefei, Anhui, in 2009, and the Ph.D. degree in control science and engineering from Jiangnan University, Wuxi, Jiangsu, in 2016. Since 2015, he has been an Associate Professor with the School of Mathematics and Computational Science, Anqing Normal University. His research interests include pattern recognition and intelligent systems, and machine learning.
\end{IEEEbiography}

\EOD

\end{document}